\documentclass[journal]{IEEEtran}
\usepackage{cite}
\usepackage{algorithmicx}
\usepackage{textcomp}

\usepackage{booktabs}
\usepackage{comment}
\usepackage[printonlyused]{acronym}

\usepackage{algorithm}
\usepackage{algpseudocode}
\usepackage{makecell}
\usepackage{subfig}
\usepackage{multirow}
\usepackage{graphicx}

\usepackage{hyperref}
\usepackage{url}
\usepackage{subfig}
\usepackage{booktabs}
\usepackage{threeparttable}
\usepackage{adjustbox}
\usepackage[normalem]{ulem}
\usepackage{multirow}
\usepackage{textcomp}

\usepackage[fleqn]{amsmath}
\usepackage{amsmath,amssymb,amsfonts}
\usepackage{pifont}
\usepackage{mathtools}

\usepackage{todonotes}
\usepackage{lipsum}
\useunder{\uline}{\ul}{}

\let\labelindent\relax
\usepackage{enumitem}

\usepackage[en-US]{datetime2}

\makeatletter
\newcommand{\monthyear}{%
  \DTMenglishmonthname{\@dtm@month}, \@dtm@year
}
\makeatother

\makeatletter
\newcommand{\thisyear}{%
 \@dtm@year
}
\makeatother

\usepackage{bbding}
\usepackage{cite}

\def\BibTeX{{\rm B\kern-.05em{\sc i\kern-.025em b}\kern-.08em
    T\kern-.1667em\lower.7ex\hbox{E}\kern-.125emX}}
\definecolor{applegreen}{rgb}{0.55, 0.71, 0.0}

\def\BibTeX{{\rm B\kern-.05em{\sc i\kern-.025em b}\kern-.08em
    T\kern-.1667em\lower.7ex\hbox{E}\kern-.125emX}}
\begin{document}
\title{IMITATE: Clinical Prior Guided Hierarchical  Vision-Language Pre-training}
\author{Che Liu,~
        Sibo Cheng,~
        Miaojing Shi,~
        Anand Shah,~
        Wenjia Bai,~
        Rossella Arcucci
\thanks{Manuscript received \today.}
\thanks{Che Liu and Rossella Arcucci are with Department of Earth Science and Engineering, Imperial College London, SW7 2AZ, UK.}
\thanks{Sibo Cheng is with CEREA, \'{E}cole des Ponts and EDF R\&D, \^Ile-de-France, France.}
\thanks{Wenjia Bai is jointly with Department of Computing and Department of Brain Science, Imperial College London, SW7 2AZ, UK.}
\thanks{Anand Shah is with Department of Infectious Disease Epidemiology, Imperial College London, SW7 2AZ, UK.}
\thanks{Anand Shah is also with Royal Brompton and Harefield Hospital.}
\thanks{Che Liu, Wenjia Bai and Rossella Arcucci are also with Data Science Institute,
Imperial College London, SW7 2AZ, UK.
}
\thanks{
Miaojin Shi is  with Tongji University.
(Corresponding author: Che Liu, email: che.liu21@imperial.ac.uk, Miaojing Shi, email: mshi@tongji.edu.cn,
Wenjia Bai, email: w.bai@imperial.ac.uk).}
}

\maketitle  
\begin{abstract}
In the field of medical Vision-Language Pre-training (VLP), significant efforts have been devoted to deriving text and image features from both clinical reports and associated medical images. However, most existing methods may have overlooked the opportunity in leveraging the inherent hierarchical structure of clinical reports, which are generally split into `findings' for descriptive content and `impressions' for conclusive observation. Instead of utilizing this rich, structured format, current medical VLP approaches often simplify the report into either a unified entity or fragmented tokens. In this work, we propose a novel clinical prior guided VLP framework named IMITATE to learn the structure information from medical reports with hierarchical vision-language alignment. The framework derives multi-level visual features from the chest X-ray (CXR) images and separately aligns these features with the descriptive and the conclusive text encoded in the hierarchical medical report. Furthermore, a new clinical-informed contrastive loss is introduced for cross-modal learning, which accounts for clinical prior knowledge in formulating sample correlations in contrastive learning. The proposed model, IMITATE, outperforms baseline VLP methods across six different datasets, spanning five medical imaging downstream tasks. Comprehensive experimental results highlight the advantages of integrating the hierarchical structure of medical reports for vision-language alignment.
\end{abstract}

\begin{IEEEkeywords}
Self-supervised Learning, Vision-Language Pre-training, Chest X-ray Image Analysis
\end{IEEEkeywords}

\section{Introduction}
\label{sec:intro}
\begin{figure}[ht!]
    \centering
    \includegraphics[width=1\linewidth, height=5.5cm]{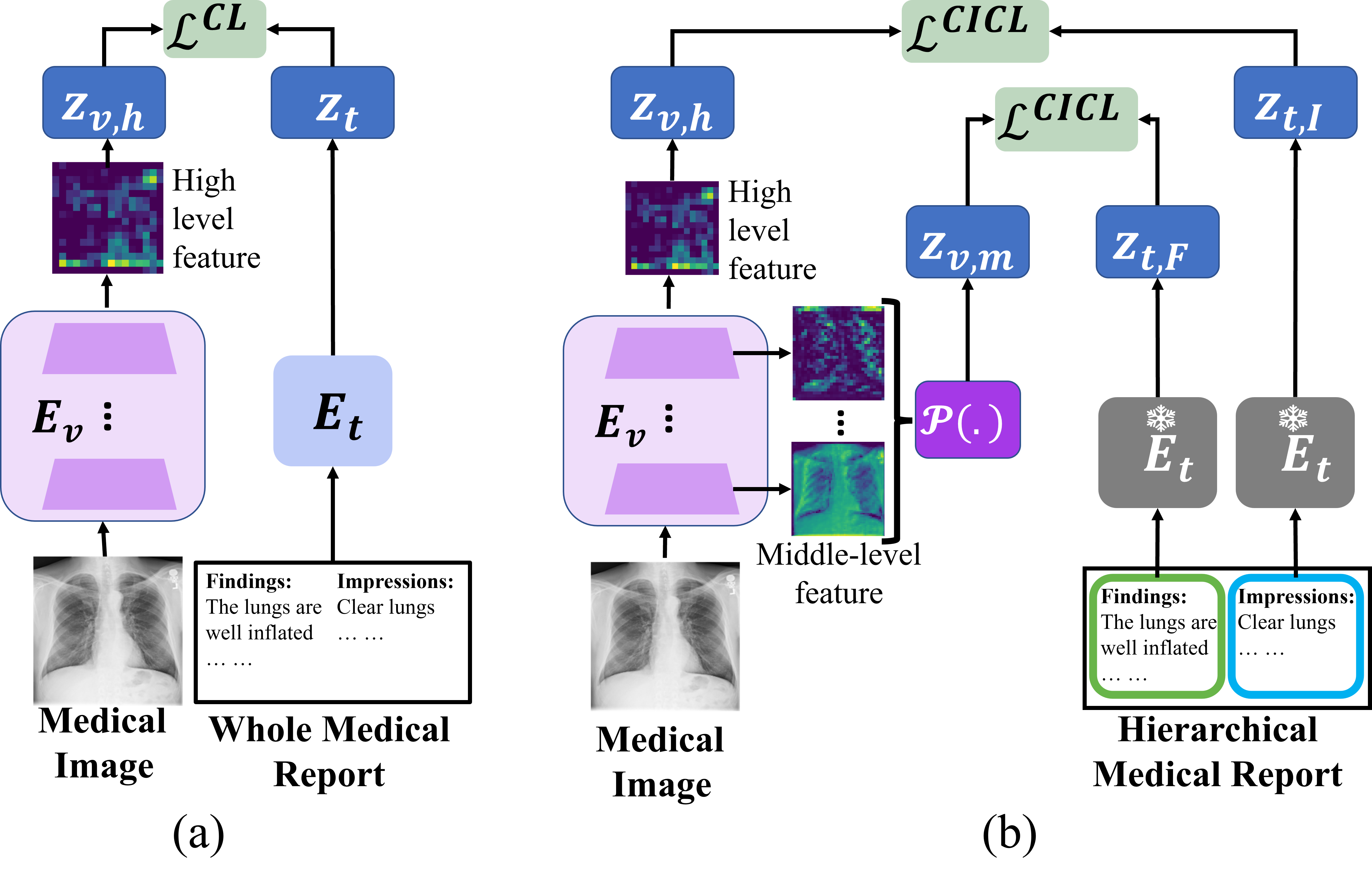}
    \caption{Architecture comparison between conventional VLP methods and the proposed method, IMITATE.
    (a) Conventional VLP approaches~\cite{convirt,clip,huang2021gloria,mgca} align the high-level visual feature with the entire medical report via a classic contrastive loss ($\mathcal{L}^{CL}$). (b) IMITATE leverages clinical prior knowledge to perform hierarchical alignment between multi-level visual features from medical images and descriptive and conclusive textual features from medical reports. Moreover, it utilizes a clinically-informed contrastive loss ($\mathcal{L}^{CICL}$), which takes into account clinical correlations among different image-report pairs. $E_v$ and $E_t$ denotes the vision and text encoders respectively. \SnowflakeChevron $E_t$ denotes a frozen text encoder. $\mathcal{P}(\cdot)$ indicates the hierarchical aggregation block. }
    \label{fig1:}
\end{figure}

Self-supervised learning has made significant progress in representation learning from a single modality such as image or text~\cite{esteva2021deep,chai2021deep,devlin2018bert,denize2023similarity}. To link the representations between different modalities, vision-language pre-training (VLP) has been introduced to align the vision and language content typically in large datasets~\cite{clip}. 
In the medical domain, as a frontline triaging and diagnosis tool, chest x-ray (CXR) scans are often accompanied by text reports as the result of the standard clinical procedure, providing a rich source of paired image-text data for VLP.
The challenge in medical VLP arises from the structure of CXR text reports, which consist of two parts, `Findings' and `Impressions'. The `Findings' section describes the image content, e.g. `the lungs are well inflated', whereas the `Impressions' section concludes the report, e.g. `clear lungs'.
Conventional VLP methods align high-level visual features with the entire medical report, without distinguishing between the descriptive and conclusive sections in the report~\cite{clip,convirt,huang2021gloria,mgca}.

To better utilise the hierarchical information in the medical report, we propose a novel clinical prior guided VLP framework, IMITATE,  
that aims to perform VLP via h\textbf{I}erachical \textbf{M}ult\textbf{I}-level con\textbf{T}r\textbf{A}s\textbf{T}ive l\textbf{E}arning.  
As depicted in Fig.~\ref{fig1:}, our framework aligns different levels of visual features separately with the descriptive and conclusive parts of the medical report. 
We hypothesize that low-level visual features embody more descriptive properties of images corresponding to the descriptive part of the report, while high-level visual features contain more semantic information corresponding to the conclusive part of the report. Apart from aligning between visual and textual features, we also align between visual features of different views to enhance the model's invariance to view variation.  
To perform the alignment, current VLP approaches perform one-to-one alignment between each image-text pair~\cite{clip,lee2022uniclip}, while ignoring the clinical similarity across different pairs. This can be problematic especially in the medical domain, because different patients may share similar symptoms, which makes their imaging scans or medical reports similar. We need to be cautious in defining the contrastive loss for different patients. 
To address this issue, we introduce a new alignment loss function named Clinical-Informed Contrastive Loss (CICL). This function integrates clinical correlations among patients into the alignment loss mechanism. Unlike traditional approaches that use a binary affinity matrix as the target \cite{clip,convirt,mgca}, CICL constructs the affinity matrix based on the similarity among different image-report pairs.

We compare the proposed method with the state-of-the-art (SOTA) VLP approaches and evaluate them on a variety of downstream tasks, including supervised medical image classification, semantic segmentation, object detection and zero-shot image classification. We show that our method significantly outperforms the SOTA methods on six public CXR datasets.
 
Overall, the contributions of this work are three-fold:
\begin{enumerate}
    \item We address the alignment challenge in medical VLP via the hierarchical alignment between multi-level visual features from medical images and the descriptive and conclusive textual features from medical reports. 
    \item We propose a new clinical-informed contrastive loss for visual-textual alignment, which incorporates the similarity among different patients into the alignment.  
    \item We achieve SOTA results across 6 datasets encompassing five distinct medical image tasks. Notably, IMITATE stands out by attaining superior performance on RSNA segmentation even with just 1\% of data for fine-tuning. This accomplishment surpasses the performance of other baseline methods that require 100\% data for fine-tuning.
\end{enumerate}

\begin{figure*}[h!]
\centering
\subfloat[Framework of IMITATE. \label{frame}]
{\includegraphics[width=0.95\linewidth,height=4.5cm]{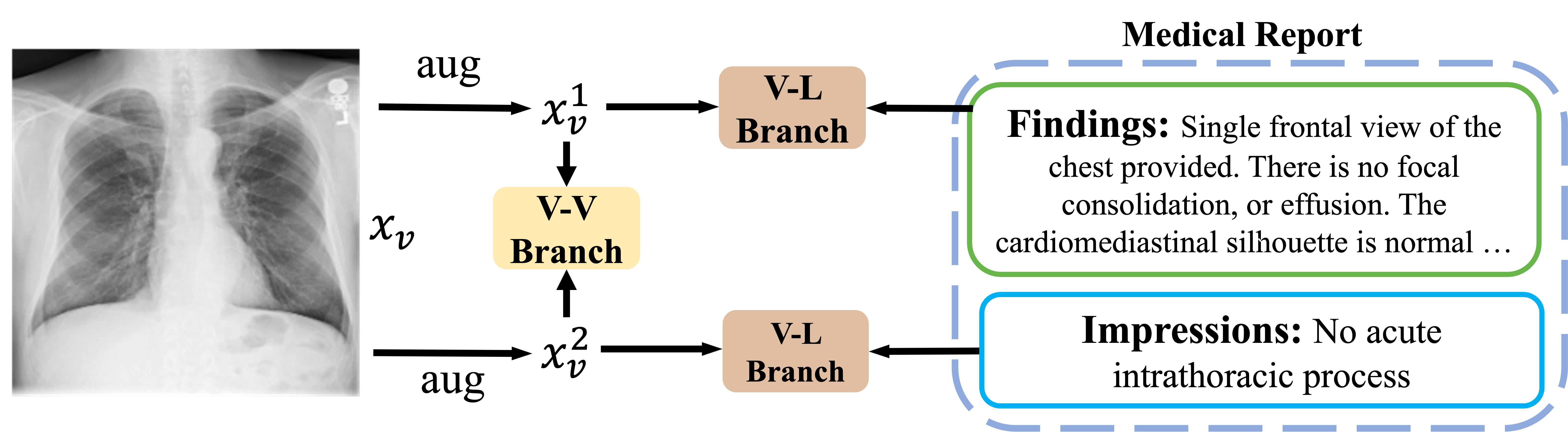}}
\vspace{-2mm}
\\
\subfloat[Vision-to-Vision (V-V) Branch. \label{v-v}]
{\includegraphics[width=0.47\linewidth,height=4.5cm]{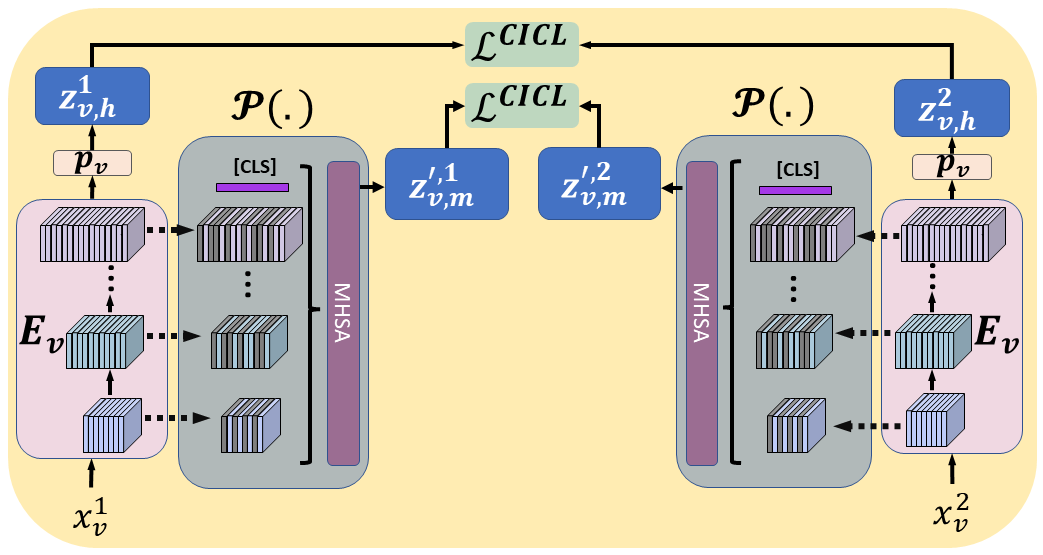}}
\hspace{3mm}
\subfloat[Vision-to-Language (V-L) Branch. \label{v-l}]
{\includegraphics[width=0.47\linewidth,height=4.5cm]{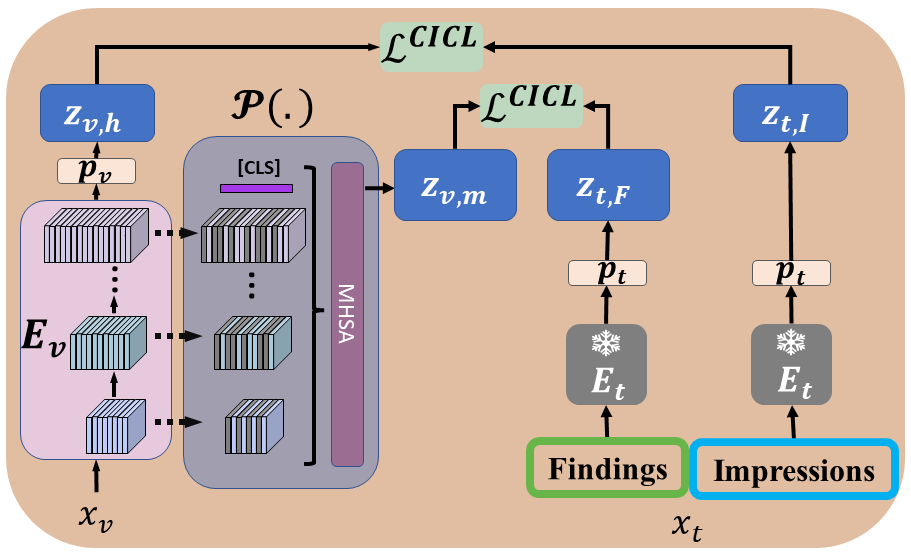}}
\caption{Overview of the proposed framework. (a) Each image is augmented to two different views ($x_{v}^{1}, x_{v}^{2}$) and provided as input to a vision-to-language (V-L) alignment branch and a vision-to-vision (V-V) alignment branch.
(b) The V-V branch aligns the visual features of two augmented views. MHSA indicates the multi-head self-attention mechanism. $p_v$ denotes a non-linear projector for visual features. CLS indicates the special token to aggregate multi-level visual features. The dashed black line indicates the feature channel dropping mechanism.
(c) The V-L branch aligns different levels of visual features to the text features from the Findings and Impressions sections of the report. $p_t$ denotes a non-linear projector for textual features. The [CLS] token serves to aggregate multi-level visual features to $z_{v,m}$ and facilitate hierarchical alignment between visual features and $z_{t, F}$.  \SnowflakeChevron $E_t$ denotes a frozen pre-trained language model.
}
\label{fig2: framework}
\vspace{-5mm}
\end{figure*}

%

\section{Related Work}
\label{sec: related work}

\subsection{General Vision-Language Pre-training}
\noindent\textbf{Joint training of Vision and Language Models\hspace{2mm}} In an effort to overcome the limitations of single-modality learning and better utilize the relationship between image and text data, VLP has been introduced~\cite{hossain2019comprehensive,baltruvsaitis2018multimodal,lu2019vilbert}, which learns and aligns representations from visual and textual input. Recent methods such as CLIP~\cite{clip}, ALIGN~\cite{align}, Florence~\cite{florence}, LiT~\cite{lit} and ALBEF~\cite{li2021align} have shown significant progress in improving the representation learning for natural images and languages, although they require substantial training data and computational resources~\cite{clip,schuhmann2021laion,schuhmann2022laion}.
Alternative methods have been proposed to reduce the training cost or data need in VLP. For instance, BeiT3~\cite{beit3} employs masked inputs and multi-modal data reconstruction to avoid the comparison of image-text pairs. A-FLIP~\cite{aflip} utilizes partially masked images as input to decrease the computational cost. SLIP~\cite{slip} introduces an additional image contrastive branch to improve CLIP~\cite{clip}.

The utilization of multi-level text in VLP can be found in PyramidCLIP~\cite{gao2022pyramidclip}.
However, PyramidCLIP does not fully embody self-supervised learning, as it employs an additional object detector to extract regional visual features during the VLP stage. This detector is trained on annotated data complete with bounding box labels.
Also, PyramidCLIP uses randomly cropped images to align with the text, which, according to ~\cite{van2023exploring}, can be inappropriate for medical imaging where anatomical correspondence is important when matching between the image and text. 

\noindent\textbf{Frozen Language Model in VLP\hspace{2mm}}
VLP requires tremendous computational resources which can be prohibitively expensive. To address this issue, ~\cite{taleb2021multimodal,yang2022zero} freeze the language model for VLP and achieve competitive results on the visual question answering task. Furthermore, a recent work of~\cite{li2023blip} freezes both the language and vision models and designs an additional trainable block to align visual and language embeddings.
These methods demonstrate the potential of using a frozen language model in VLP. In this paper, we investigate the frozen language model in medical VLP. 

\subsection{Medical Vision-Language Pre-training}
Research in medical VLP is limited due to the complexity of the medical reports and the scarcity of large-scale medical image-text datasets.
ConVIRT~\cite{convirt} pioneered VLP within the CXR domain, leveraging a bidirectional contrastive loss to align image-text pairs. GLoRIA~\cite{huang2021gloria} proposed a global-local VLP technique that seeks alignment between image regions and individual text tokens, fostering enhanced representation learning. MGCA~\cite{mgca} adopted a crafted prototype-level alignment strategy to align semantic relationships between CXR images and their associated reports. Notably, these methodologies~\cite{huang2021gloria,mgca} attempt to align comprehensive medical reports to high-level visual features, while fragmenting the report into word-level tokens for local alignment. However, this token-level alignment might compromise the medical context, potentially leading to misalignments. For instance, within the `Impressions' section, terms like `compatible' or `acute' lack direct visual correlates, rendering local alignment potentially ambiguous.

While MedKLIP~\cite{medklip} and KAD~\cite{kad} utilize domain-specific knowledge from external datasets to enhance textual information extraction, one might argue about their dependence on these external resources for vision-language alignment.
Furthermore, Med-UniC~\cite{medunic}, which integrates multi-lingual medical data for VLP, aims to analyze and mitigate language biases originating from different communities in the realm of medical VLP.
MRM~\cite{zhouadvancing} shifts from the alignment task to a reconstruction task that uses masked visual and textual tokens.

These studies enhance the performance of medical VLP in various ways; however, they do not account for a clear distinction between the descriptive and conclusive segments within a medical report. Moreover, the potential similarity inherent in medical data is overlooked during the execution of vision-language alignment, which in turn adversely affects the cross-modal representation learning.

\section{Method}

\subsection{Overview}
Our framework overview is illustrated in  Fig.~\ref{fig2: framework}. It is composed of a ResNet50~\cite{resnet} vision encoder, denoted by $E_v$, followed by a hierarchical aggregation block $\mathcal{P}(\cdot)$; and a BioClinicalBERT~\cite{alsentzer2019publicly} text encoder denoted by $E_t$ (as described in Sec.~\ref{sec: ext feature}). 
During the pre-training stage, we update the parameters of \(E_v\) and \(\mathcal{P}(\cdot)\), while \(E_t\) is frozen to prevent disturbance from the language side and also minimize the training expense. To align the visual and text features, we also train visual and text projectors to project the visual and text features to the same dimensions.
The proposed framework aims to both optimize the vision-to-vision (V-V) branch between visual features of different views and the vision-to-language (V-L) branch between visual and textual features.

Given a set of $N$ image-report pairs  $\mathcal{X} = \{ (x_{v,i}, x_{t,i})\}_{i = 1}^N$, where $(x_{v,i}, x_{t,i})$ denote the paired image and text, each report $x_{t,i}$ includes two parts: `Findings' and `Impressions', such that $x_{t,i}$ can be split into $x_{F,i}$  and $x_{I,i}$. Therefore, each image-report pair can be further represented by $(x_{v,i}, x_{F,i}, x_{I,i})$. For simplicity, unless needed, we omit the subscript $i$ in later text. 

\subsection{Semantic Difference in Hierarchical Medical Report}

\begin{table}[ht!]
\caption{An exemplar CXR report.}
\label{tab: report}
\centering
\scalebox{1.2}{
\begin{tabular}{lll}
\multicolumn{3}{l}{\textbf{INDICATION:} \text{\textit
{Patient Name} with cough / acute process?}} \\ \hline
\multicolumn{3}{l}{\textbf{FINDINGS:} Single frontal view of the chest provided.} \\
\multicolumn{3}{l}{The cardiomediastinal silhouette is normal.} \\
\multicolumn{3}{c}{\textbf{... ...}} \\ 
\multicolumn{3}{l}{No free air below the right hemidiaphragm is seen.} \\\hline
\multicolumn{3}{l}{\textbf{IMPRESSIONS:} No acute intrathoracic process.}
\end{tabular}
}
\end{table}

\begin{figure*}[t!]
    \centering
    \includegraphics[width=1\linewidth]{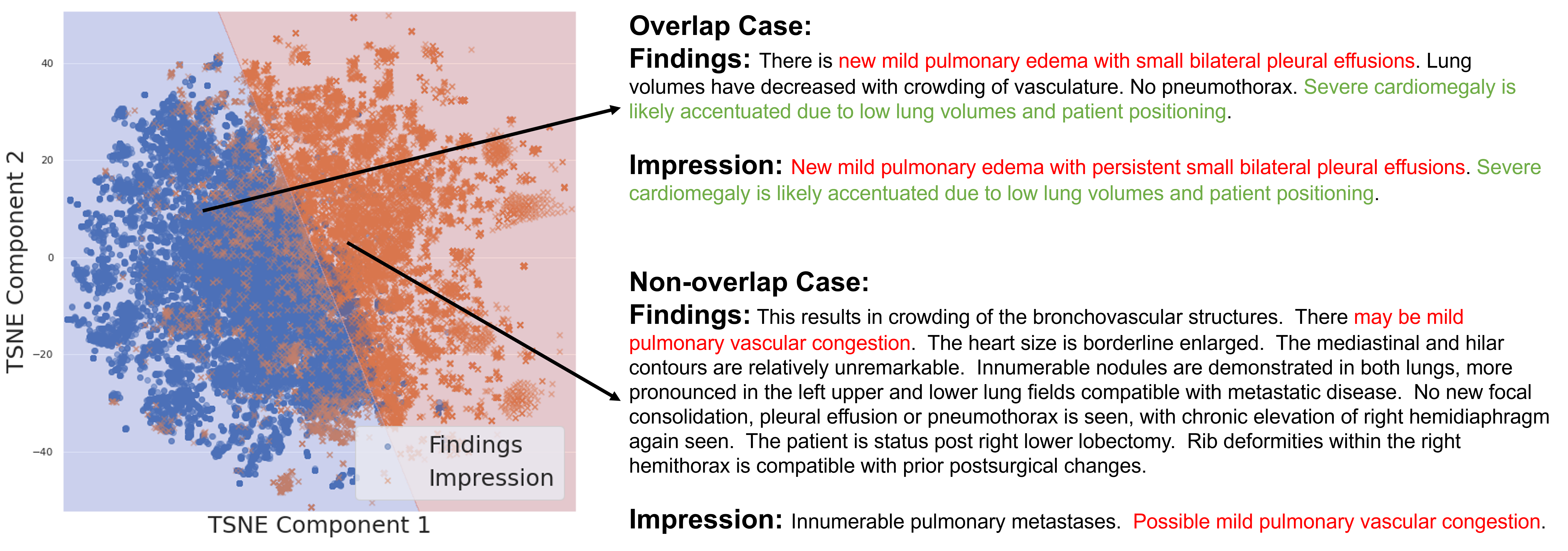}
    \caption{\textbf{Left:} 2D TSNE visualization of text embedding from `Findings' and `Impression'. We also plot the decision boundary of two parts with KMeans. \textbf{Right:} Two reports samples from the overlap and non-overlap area respectively. Rephrased sentence are highlighted with the same color.}
    \label{fig: report tsne}
\end{figure*}

As shown in Tab \ref{tab: report}, the `Findings' section contains descriptive content, whereas the `Impressions' section offers a conclusive remark.
To examine the distribution in the latent space, we randomly selected 40,000 reports from MIMIC-CXR \cite{johnson2019mimic} and extracted the text embeddings for the `Findings' and `Impressions' sections using a pre-trained text encoder.
We then visualized the latent features of `Findings' and `Impression' from these 40,000 samples in Figure \ref{fig: report tsne}. In the left part of Figure \ref{fig: report tsne}, we employed TSNE \cite{van2008visualizing} to reduce the feature dimensions and depict them in 2D space. 
For enhanced visualization, the decision boundary was plotted with KMeans \cite{macqueen1967some} in the left part of Fig \ref{fig: report tsne}. 
We considered these two report sections as distinct clusters and applied KMeans for clustering, subsequently delineating the boundary between the two clusters. 
It is apparent that the latent features of `Findings' and `Impression' are mostly distinct in the 2D latent space.

Nonetheless, areas of overlap remain. To delve deeper, we extracted examples from both overlapping and distinct areas, showcased in the right panel of Fig \ref{fig: report tsne}. 
In an overlapping example, the `Findings' section is less detailed compared to an example from a non-overlapping area. Moreover, the `Impression' in the overlapping example nearly duplicates the `Findings' content, with minimal alterations. 
This similarity likely brings their text embeddings closer in the latent space.
Conversely, in the non-overlapping example, the `Findings' section thoroughly describes the CXR image, encompassing various anatomical regions, while the `Impression' section summarize rather than replicates the `Findings', indicating different semantic levels of description for the same CXR image. Thus, the overlap could stem from varying radiologist writing styles.
In light of this, aligning images with entire medical reports runs the risk of introducing misalignment errors. 
To address this challenge, we devise a sophisticated hierarchical alignment strategy that both mitigates the issue of misalignment and more effectively capitalizes on the inherent hierarchical structure of medical reports.

\subsection{Hierarchical Vision-Language Alignment}
\label{sec: ext feature}
To align visual features separately to the `Findings' and `Impressions' of a medical report, 
we extract multi-level features from the medical image and develop the hierarchical alignment scheme, which includes the hierarchical feature aggregation and intermediate channel dropping mechanism to make the learning more effective and efficient.

\noindent \textbf{Hierarchical feature aggregation}\hspace{2mm}
As illustrated in Fig.\ref{fig2: framework}, the hierarchical feature aggregation mechanism, $\mathcal{P}(\cdot)$, employs multi-head attention (MHSA)~\cite{vaswani2017attention}, positional embedding~\cite{vaswani2017attention}, and a [CLS] token to aggregate diverse feature levels. This architecture is meticulously devised to enhance the understanding of the distinct sections, `Findings' and `Impression', in medical reports. The `Findings' section predominantly reflects low-level image features, e.g., intensity, texture and size, while the `Impression' encapsulates image semantics, e.g., disease and symptom manifestations~\cite{mimic3,johnson2019mimic}.

In the proposed method, multi-level visual features, extracted from the output of each ResBlock~\cite{resnet} in the vision encoder, are designated as middle-level features, aligning with the detailed nature of the `Findings' section. In contrast, the terminal output from the vision encoder, denoted as $E_v$, is treated as the high-level visual feature $z_{v,h}$, aligning with the summarization text in the `Impressions' section.

We use $(h,w,c)$ to denote the size of visual features from different levels, where $h$ and $w$ represent the height and width of the feature map, and $c$ denotes the number of channels. To standardize the shape of these features, we pool and flatten all feature maps to a size of $(16, 16, c)$. Then, we treat these pooled and flattened features as a sequence, which has a shape of $(256, c)$. $c$ is also the number of tokens of a sequence. To aggregate the sequence visual features, we concatenate a [CLS] token and a positional embedding to the features following~\cite{dosovitskiy2020image}. 
The multi-level features have channel sizes of 256, 512, 1024 and 2048 from the shallower to the deeper blocks, respectively. 
We introduce an intermediate  feature channel dropping scheme to squeeze the numbers of feature channels for them, detailed in the next subsection. Afterward, we concatenate all flattened feature maps with the positional embedding and the [CLS] token. They are embedded using MHSA similar to that in a transformer~\cite{vaswani2017attention}. To aggregate the sequence, only the [CLS] token embedding from the MHSA output layer is utilized for later alignment.  

\noindent\textbf{Intermediate feature channel dropping}\hspace{2mm}  
If the multi-level (middle-level) features are used directly for subsequent alignment via MHSA, there might be a significant increase in computational cost. This is because the computational complexity is $\mathcal{O}(N^{2})$ for MHSA~\cite{vaswani2017attention}. Here, $N$ represents the total length of the visual feature sequence from all ResBlocks\cite{resnet}. 
To alleviate this issue, We propose to randomly drop the feature channels of these middle-level features. The drop ratio is set to $0.85$ for the shallowest level and $0.9$ for the other levels. 

Above we have the multi-level visual features ready through the hierarchical feature aggregation module. Below we introduce how to get the high-level visual features, as well as the descriptive and conclusive textual features. We organize the alignment in two branches, i.e., V-L and V-V branches.  

\noindent\textbf{V-L branch}\hspace{2mm} In the V-L branch, as depicted in Fig.~\ref{v-l}, given the `Findings' and `Impressions' sections of the text report, the text encoder $E_t$ is utilized to extract text embedding $(z_{t,F}, z_{t,I})$ correspondingly. 
Given the medical image, the last layer output of the vision encoder $E_v$ is considered as the high-level visual feature $z_{v, h}$, and the features from intermediate layers of $E_v$ are aggregated and projected to the multi-level visual feature $z_{v, m}$ using the above mentioned hierarchical aggregation block $\mathcal{P}(\cdot)$. 
Hence, one image-report pair $(x_{v}, x_{F}, x_{I})$ is embedded to $(z_{v, m}, z_{v, h}, z_{t,F}, z_{t,I})$, where the subscripts $v$ and $t$ denote visual and textual respectively.

\noindent\textbf{V-V branch}\hspace{2mm}
In the V-V branch, as illustrated in Fig.~\ref{v-v}, we augment each medical image into two different views, \(x_{v}^{1}\) and \(x_{v}^{2}\), to promote the learning of invariant image features. The augmentation is implemented following the methods in~\cite{huang2021gloria,mgca}.

Then, the two augmented images are embedded to $(z_{v, m}^{1}, z_{v, h}^{1}, z_{v, m}^{2}, z_{v, h}^{2})$ via $E_v$.

Finally, each image-report pair $(x_{v}, x_{F}, x_{I})$ is embedded into $(z_{v,m}^{1}, z_{v,h}^{1}, z_{v,m}^{2}, z_{v,h}^{2}, z_{t,F}, z_{t,I})$. 
Next, we perform hierarchical alignment between these visual and textual features.

\noindent \textbf{Hierarchical feature alignment}
\label{cicl}
After obtaining all visual and textual features, 
we perform the hierarchical alignment by using the clinical-informed contrastive loss (CICL), which includes three parts: 
\begin{itemize}
    \item In the V-L branch, we minimize the CICL between  $(z_{v,h}^{1}$ and $z_{t,I})$,  $(z_{v,h}^{2}$ and $z_{t,I})$ to align the high-level visual feature from the image and the conclusive  textual feature from the report. 
    \item In the V-L branch, we minimize the CICL between $(z_{v,m}^{1}$ and $z_{t,F})$, $(z_{v,m}^{2}$ and $z_{t,F})$ to align the multi-level visual feature from the image and the descriptive textual feature from the report. 
    \item In the V-V branch, we minimize the CICL between $(z_{v,m}^{1}$ and $z_{v,m}^{2})$, $(z_{v,h}^{1}$ and $z_{v,h}^{2})$ to align the visual features across multiple views. 
\end{itemize}

\subsection{Clinical-Informed Contrastive loss}

Contrastive loss is the most commonly used optimization strategy for VLP~\cite{clip,convirt,huang2021gloria}. Given the visual and textual features, $z_v$ and $z_t$, the contrastive loss is defined as below:  

\begin{equation}
\label{eq: ori cl loss}
\mathcal{L}^{CL}=\sum_{i=1}^B \Bigg(-\log \frac{\exp \left({sim}\left(z_{v}^{i}, z_{t}^{i}\right)/\tau \right)}{\sum_{j=1}^B \exp \left(sim\left(z_{v}^{i}, z_{t}^{j}\right)/\tau \right)} \Bigg),
\end{equation}

where $B$ denotes the size of a mini-batch, $i,j$ denote the sample index in the mini-batch.
\(\tau\) indicates the temperature parameter; we set it to 0.07 following \cite{huang2021gloria,mgca}.
Two trainable non-linear projectors $p_v$ and $p_z$ are utilized to map the visual embedding $z_v^{i}$ and the text embedding $z_{t}^{i}$ into the same latent space for the alignment.  
In our case, 
we use the $\{z^i_1, z^j_2\}$ to represent visual or textual embedded features:
\begin{align*}
    \{z^i_1, z^j_2\} \in \{z_{v,m}^{1,i}, z_{v,h}^{1,i}, z_{v,m}^{2,i}, z_{v,h}^{2,i}, z_{t,F}^{i}, z_{t,I}^{i}\} \times \\
    \{z_{v,m}^{1,j}, z_{v,h}^{1,j}, z_{v,m}^{2,j}, z_{v,h}^{2,j}, z_{t,F}^{j}, z_{t,I}^{j}\} ,
\end{align*}
The similarity of two components $sim(z^i_1, z^j_2)$ is computed as,

\begin{equation}
    sim(z^i_1, z^j_2) = p_1(z_1)^{T} p_2(z_2). 
\end{equation}
where $\{p_1,p_2\} \in \{ p_v,p_z\}^2$, representing visual or text projectors. 

In contrastive learning, sample $i$ and sample $j$ denote two patients. Although they are two different patients, they may present similar symptoms and in this case, it implies the textual descriptions of their reports will exhibit higher similarity, so as are their CXR images. Conventional  approaches~\cite{convirt,slip,huang2021gloria,mgca} do not consider this factor in the alignment and assume that two different patients have dissimilar visual and textual features. 
Our idea is to leverage the clinical information in the medical report to adaptively adjust the contrastive loss.
To do so, we first compute an empirical report correlation matrix $R \in \mathbb{R}^{B \times B}$:

\begin{equation}
R_{i,j} = \frac{\sum_{k=1}^B (z_{t}^{i, (k)} - \overline{z}_{t}^{i}) (z_{t}^{j, (k)} - \overline{z}_{t}^{j})}{\sqrt{\sum_{k=1}^B (z_{t}^{i, (k)} - \overline{z}_{t}^{i})^2} \sqrt{\sum_{k=1}^B (z_{t}^{j, (k)} - \overline{z}_{t}^{j})^2}},
\end{equation}
where $z_{t}^{i, (k)}$ denotes the $k^{\textrm{th}}$ element of the latent code $z_{t,F}^{i}$ or $z_{t,I}^{i}$. $\overline{z}_{t}^{i} = \sum_{k=1}^B z_{t}^{i, (k)}/B$ is the averaged vector. To avoid ill-defined empirical correlation matrix and centralize the correlation coefficients, we smooth $R_{i,j}$ as
\begin{align}
\label{eq:smooth}
    R^{i,j}_{\text{smooth}} =
    \begin{cases}
        1 & \text{if } i = j \\
        1-\exp^{(-\lambda R^{i,j})} & \text{otherwise}
    \end{cases},
\end{align}
where $\lambda$ denotes a regularization coefficient, empirically set to 0.2. This hyperparameter is extensive ablated in Sec.~\ref{abla: cicl} and Fig.~\ref{fig: lambda analysis}.

The key component of $\mathcal{L}^{CICL}$ consists of approximating the smoothed text correlation based on the similarity of visual and textual components. We introduce the clinical-informed contrastive loss as

\begin{align*}
\mathcal{L}^{CICL}&=\sum_{i=1}^B \sum_{j=1}^B \mathcal{L}_{CrossEntropy} ({sim}(z_{1}^{i}, z_{2}^{j}),\hspace{1mm}  R_\text{smooth}^{i,j}).
\end{align*}

\subsection{Total Loss}
As mentioned in Sec.~\ref{cicl}, $\mathcal{L}^{CICL}$ is applied different terms in the V-V and V-L branches, leading to the final loss function,

\begin{align*}
    \mathcal{L}_{total} = \mathcal{L}^{CICL}(z_{v,h}^{1}, z_{I}) + \mathcal{L}^{CICL}(z_{v,m}^{1}, z_{F})\\
    + \mathcal{L}^{CICL}(z_{v,h}^{2}, z_{I}) + \mathcal{L}^{CICL}(z_{v,m}^{2}, z_{F})\\
    + \mathcal{L}^{CICL}(z_{v,h}^{1}, z_{v,h}^{2}) + \mathcal{L}^{CICL}(z_{v,m}^{1},z_{v,m}^{2}).
\end{align*}

\section{Experiments and Analysis}

\subsection{Vision-Language Pre-training Configuration}
\noindent\textbf{Dataset}\hspace{2mm} Our method, IMITATE, is pre-trained on the MIMIC-CXR dataset~\cite{johnson2019mimic,johnson2019mimicjpg}. The preprocessing of this dataset adheres to practice described in~\cite{convirt,huang2021gloria,mgca}, including image resizing, pixel value normalization, and text tokenization. To refine the dataset further, lateral views and reports comprising fewer than three tokens were excluded, resulting in a pre-training dataset with $213,384$ image-text pairings for MIMIC-CXR~\cite{johnson2019mimicjpg}.

\noindent\textbf{Implementation}\hspace{2mm} The original CXR images from the MIMIC-CXR dataset~\cite{johnson2019mimicjpg} are resized to $256\times 256$ and randomly cropped to $224\times 224$, following the procedure in~\cite{convirt,huang2021gloria,mgca}. All images are normalized to the range $[0,1]$. For data augmentation during pre-training, we apply horizontal flip, random rotation in the range $[0^{\circ},180^{\circ}]$, and auto contrast using the PyTorch vision library\footnote[1]{https://pytorch.org/vision/stable/transforms.htmls}.
We employ BioClinicalBERT~\cite{alsentzer2019publicly} to derive text embeddings from medical reports. 
To conduct a fair comparison with existing methods~\cite{convirt, huang2021gloria, mgca, kad, medklip}, we utilize the same ResNet50~\cite{resnet} backbone as our vision encoder. The weights are initialized with ImageNet pre-trained weights\footnote{https://github.com/huggingface/pytorch-image-models/tree/main} following \cite{mgca, medklip, kad}.
Adhering to practices outlined in~\cite{mgca,huang2021gloria}, the proposed model is pre-trained for 50 epochs. This training utilizes an early stopping strategy and is conducted on 16 A100-40GB GPUs parallel, each handling a batch size of 128. For optimization, we employ the AdamW optimizer, configuring a learning rate of $4e^{-5}$ and a weight decay of $5e^{-2}$. Throughout this phase, a combination of a linear warm-up and a cosine annealing scheduler is employed~\cite{loshchilov2016sgdr}.

\subsection{Downstream Tasks}
We evaluate our framework on five downstream tasks:

\noindent \textbf{Medical Image Classification}
We evaluate on the CheXpert~\cite{irvin2019chexpert}, RSNA~\cite{rsna}, COVIDx~\cite{wang2020covid}, and ChestX-ray14 datasets~\cite{wang2017chestx}. The latter consists of 112,120 X-ray images annotated for 14 diseases. Our metrics include macro-averaged AUC scores for CheXpert, RSNA, and ChestX-ray14, as well as accuracy for COVIDx.

Following prior studies~\cite{huang2021gloria,convirt,mgca,medklip,kad}, we implement linear classification on CheXpert~\cite{irvin2019chexpert}, RSNA~\cite{rsna}, and COVIDx~\cite{wang2020covid}. We update a randomly initialized linear layer while keeping the backbone frozen. For consistency, we employ the official test set partition from~\cite{medklip,kad,wang2017chestx}.

In our linear classification approach, fine-tuning is conducted over 50 epochs with a learning rate of 5e-4, a batch size of 8, and the AdamW optimizer parameters: $\beta_{1} = 0.9$ and $\beta_{2} = 0.999$.

For the ChestX-ray14 dataset~\cite{wang2017chestx}, following the experimental configuration from \cite{kad}, we deploy fine-tuned classification and update all model parameters. Images are resized to $256 \times 256$ and undergo data augmentation as suggested in~\cite{kad}. The AdamW optimizer is used with a learning rate of $1 \times 10^{-4}$ and a batch size of 64 for 50 epochs.

\noindent \textbf{Medical Image Semantic Segmentation} 
This task is performed on the RSNA~\cite{rsna} and the SIIM~\cite{siim} datasets, following the data preprocessing in~\cite{mgca,huang2021gloria}. Similar to~\cite{huang2021gloria,mgca}, the U-Net~\cite{unet} fine-tuning settings are adopted for segmentation. All pre-trained models are considered as frozen encoders and only the decoders of U-Net are updated during the fine-tuning. The segmentation performance is evaluated using Dice scores following~\cite{huang2021gloria,mgca,medklip}.

Following the preprocessing steps outlined in~\cite{huang2021gloria,mgca}, to generate the segmentation mask for pneumonia regions, we resize all images and masks to $512\times 512$, applying the same data augmentation techniques as in~\cite{mgca}. For fine-tuning, we use the AdamW~\cite{kingma2014adam} optimizer with a learning rate of $5\times 10^{-4}$ and weight decay of $1\times 10^{-6}$, and optimize the segmentation model using a combination $\alpha\times \textrm{FocalLoss+DiceLoss}$ with a coefficient $\alpha$ set to 10~\cite{mgca}. We fine-tune the segmentation task for 50 epochs and early stop if the loss does not decrease on the validation set for 10 steps.
We use a batch size of 16 for RSNA segmentation and 8 for SIIM segmentation.

\noindent \textbf{Medical Image Object Detection}
We conduct pneumonia detection on the RSNA dataset~\cite{rsna} and foreign objects detection on the Object-CXR dataset~\cite{obj-cxr}, adhering to the preprocessing standards outlined by~\cite{mgca}.
Consistent with~\cite{mgca}, we employ the YOLOv3~\cite{redmon2018yolov3} for the detection framework. Within this architecture, our pre-trained vision encoder acts as the backbone, and during fine-tuning, only the detection head is optimized. Evaluation metrics for the detection task are based on the Mean Average Precision (mAP) with IOU thresholds spanning from 0.4 to 0.75. 

According to~\cite{mgca}, we normalize all pixel intensity of images in RSNA Pneumonia~\cite{rsna} to range [0,1] and do not apply data augmentation during the fine-tuning stage for fair comparison. For all data fractions, the batch size is 16. We choose AdamW~\cite{kingma2014adam} as the optimizer with learning rate set to $5\times 10^{-4}$ and weight decay of $1\times 10^{-6}$. The detection model is trained for 50 epochs and early stopping is applied when the validation loss does not decrease for 10 steps. Other details follow~\cite{mgca}.

\noindent \textbf{Medical Image Zero-shot Image Classification}
Following~\cite{medklip,kad}, we execute this task utilizing RSNA and SIIM datasets~\cite{wang2017chestx}. To fairly compare with previous methods, we adopt the official test set split from~\cite{medklip,kad}. To alleviate potential biases stemming from human-crafted prompts, our positive prompts are structured as '{\textit{disease}}' and negative prompts as 'No {\textit{disease}}'. 
The original image undergoes a two-step process. Firstly, it is resized to dimensions of $256\times 256$ and then center cropped to $224\times 224$. Subsequently, all pixel values are normalized within the range of $[0,1]$, following~\cite{medklip,kad}. The resulting resized image is then passed through an image encoder to generate an image embedding . Concurrently, the prompts are inputted into a text encoder to obtain a text embeddings. To evaluate the classification, we measure the cosine similarity between the image and text embeddings for each prompt associated with a specific class. Our results are reported as the macro average of AUC, F1, and ACC scores across the spectrum of all diseases. 

\noindent \textbf{Medical Image-text Retrieval} For the image-text retrieval task, we adopt the method used in GLoRIA \cite{huang2021gloria}, unlike other methods that do not include this task.
We evaluate our representation learning framework using the CheXpert $5 \times 200$ dataset. The process involves using an image as an input query to retrieve target reports by computing the similarity between the query image and all candidate reports using the learned representations. Precision is measured with the Precision@ $K$ metric, determining the accuracy of the top $K$ retrieved reports, ensuring they match the category of the query image.

For all downstream tasks, we use only the visual features from the vision encoder \(E_v\) and omit the \(P(\cdot)\) module to ensure a fair comparison with other methods. However, in the zero-shot image classification and image-text retrieval task, we use both the vision encoder \(E_v\) and the text encoder \(E_t\), along with the vision projector \(p_v\) and the text projector \(p_z\), to project visual and textual embeddings to the same number of dimensions.

All data split information and train/valid/test set details are in Tab.~\ref{tab:split}. For all downstream tasks, except zero-shot classification and image-text retrieval, we train with ${1\%, 10\%, 100\%}$ of the training set. 

\begin{table}[h]
\centering
\caption{Data Split Details, `/' indicates that no training/validation data is required in the zero-shot classification task.}
\label{tab:split}
\scalebox{0.85}{
\begin{tabular}{crrrrr}
Task & Dataset & Split & Train & Valid & Test \\ \cline{1-6} 
\multicolumn{1}{c}{\multirow{3}{*}{\begin{tabular}[c]{@{}c@{}}Linear \\ Classification\end{tabular}}} & CheXpert~\cite{irvin2019chexpert} & ~\cite{irvin2019chexpert} & 186,027  & 5,000 & 202 \\
\multicolumn{1}{c}{} & RSNA~\cite{rsna} & ~\cite{mgca,rsna}& \multicolumn{1}{c}{16,010} & \multicolumn{1}{c}{5,337} & \multicolumn{1}{c}{5,337} \\
\multicolumn{1}{c}{} & COVIDx~\cite{wang2020covid} & ~\cite{mgca,wang2020covid} & \multicolumn{1}{c}{23988} & \multicolumn{1}{c}{5998} & \multicolumn{1}{c}{400} \\ \midrule
\multirow{2}{*}{\begin{tabular}
[c]{@{}c@{}}Fine-tuned \\ Classification\end{tabular}} & \multicolumn{1}{r}{ChestX-ray14~\cite{wang2017chestx}} &  \cite{kad} & \multicolumn{1}{c}{77,872} & \multicolumn{1}{c}{8,652} & \multicolumn{1}{c}{25,596} \\
 \\ \midrule
\multirow{2}{*}{\begin{tabular}
[c]{@{}c@{}}Semantic \\ Segmentation\end{tabular}} & \multicolumn{1}{r}{RSNA~\cite{rsna}} &  \cite{huang2021gloria,mgca} & \multicolumn{1}{c}{16,010} & \multicolumn{1}{c}{5,337} & \multicolumn{1}{c}{5,337} \\
 & \multicolumn{1}{r}{SIIM~\cite{siim}} & \multicolumn{1}{c}{\cite{huang2021gloria,mgca}} & \multicolumn{1}{c}{8,433} & \multicolumn{1}{c}{1,807} & \multicolumn{1}{c}{1,807} \\ \midrule
\multirow{2}{*}{\begin{tabular}
[c]{@{}c@{}}Object \\ Detection\end{tabular}} & \multicolumn{1}{r}{RSNA~\cite{rsna}} &  \cite{huang2021gloria,mgca} & \multicolumn{1}{c}{16,010} & 5,337 & 5,337 \\
 & \multicolumn{1}{r}{Object-CXR~\cite{obj-cxr}} & \multicolumn{1}{r}{\cite{mgca}} & \multicolumn{1}{c}{6,400} & \multicolumn{1}{c}{1,600} & \multicolumn{1}{c}{1,000} \\ \midrule
\multirow{2}{*}{\begin{tabular}
[c]{@{}c@{}}Zero-shot \\ Classification\end{tabular}} & \multicolumn{1}{r}{RSNA~\cite{rsna}} &  \cite{medklip} & \multicolumn{1}{c}{/} & \multicolumn{1}{c}{/} & \multicolumn{1}{c}{5,337} \\
 & \multicolumn{1}{r}{SIIM~\cite{siim}} & \multicolumn{1}{r}{\cite{medklip}} & \multicolumn{1}{c}{/} & \multicolumn{1}{c}{/} & \multicolumn{1}{c}{1,807} \\
 \midrule
\multirow{2}{*}{\begin{tabular}
[c]{@{}c@{}}{Image-text} \\ {Retrieval}\end{tabular}} & \multicolumn{1}{r}{{CheXpert $5\times200$}~\cite{huang2021gloria}} &  \cite{huang2021gloria} & \multicolumn{1}{c}{/} & \multicolumn{1}{c}{/} & \multicolumn{1}{c}{{1,000}} \\
 \\ \midrule
\end{tabular}
}
\end{table}

\subsection{Results}
In this section, we evaluate the performance of our method on five downstream tasks regarding medical images, in comparison to 8 SOTA medical VLP methods.

\subsubsection{Medical Image Classification}
To evaluate the quality of visual representation from IMITATE, we implement medical image classification on four CXR image datasets: CheXpert~\cite{irvin2019chexpert}, RSNA~\cite{rsna}, COVIDx~\cite{covid}, and ChestX-ray14~\cite{wang2017chestx}. Results in Tab.\ref{tab: ft cls} and Tab.\ref{tab: res finetune} demonstrate that IMITATE consistently outperforms other SOTA baselines across all datasets and data ratios. Notably, it surpasses MedKLIP~\cite{medklip} and KAD~\cite{kad}, both of which incorporate additional annotated data with disease-level annotations during their VLP stages. This emphasizes the efficacy of IMITATE in enhancing disease prediction.

\begin{table*}[h!]
\centering
    \caption{Linear classification results for CheXpert, RSNA, and COVIDx datasets with 1\%, 10\%, and 100\% training data. The best results are highlighted in bold.  Methods with $\star$ use extra annotated data for pre-training.}
    \scalebox{1.2}{
    
    \begin{tabular}{lccccccccc}
        \toprule[1.2pt]
         & \multicolumn{3}{c}{CheXpert (AUC)} & \multicolumn{3}{c}{RSNA (AUC)} 
        & \multicolumn{3}{c}{COVIDx (ACC)} \\
        Method & 1\% & 10\% & 100\% & 1\% & 10\% & 100\% & 1\% & 10\% & 100\% \\
        \midrule[1.2pt]
        Random Init & 56.1 & 62.6 & 65.7 & 58.9 & 69.4 & 74.1 & 50.5 & 60.3 & 70.0 \\
        ImageNet Init & 74.4 & 79.7 & 81.4 & 74.9 & 74.5 & 76.3 & 64.8 & 78.8 & 86.3 \\
        \midrule
        ConVIRT~\cite{convirt} & 85.9 & 86.8 & 87.3 & 77.4 & 80.1 & 81.3 & 72.5 & 82.5 & 92.0  \\
        GLoRIA~\cite{huang2021gloria} & 86.6 & 87.8 & 88.1 & 86.1 & 88.0 & 88.6 & 67.3 & 77.8 & 89.0 \\
        GLoRIA-MIMIC~\cite{huang2021gloria} & 87.1 & 88.7 & 88.0 & 87.0 & 89.4 & 90.2 & 66.5 & 80.5 & 88.8 \\
        MGCA~\cite{mgca} & 87.6 & 88.0 & 88.2 & 88.6 & 89.1 & 89.9 & 72.0 & 83.5 & 90.5 \\
        MRM~\cite{zhouadvancing} & 88.5 & 88.5 & 88.7 & 91.3 & 92.7 & 93.3& 66.9 & 79.3 & 90.8 \\
        MedKLIP$^{\star}$~\cite{wu2023medklip} & 86.2 & 86.5 & 87.7 & 87.3 & 88.0 & 89.3 & 74.5 & 85.2 & 90.3  \\
        \midrule
        \textbf{IMITATE} & \textbf{89.1} & \textbf{89.5} & \textbf{89.7} & \textbf{91.7} & \textbf{92.9} & \textbf{93.5} & \textbf{76.8} & \textbf{87.6} & \textbf{93.1} \\

        \bottomrule[1.2pt]
    \end{tabular}
    }
    \label{tab: ft cls}
\end{table*}

\begin{table}[t]
\caption{Evaluation of different medical VLP methods for fine-tuned image classification on the ChestX-ray14 dataset under official data split~\cite{kad,wang2017chestx}. The reported metric is macro-averaged AUC score. Best performances are highlighted in bold. Methods with $\star$ use extra annotated data for pre-training.}
\centering
\scalebox{1.3}{
\begin{tabular}{l c c c}
\toprule[1.2pt]
\multicolumn{1}{c}{} & \multicolumn{3}{c}{ChestX-ray14 (AUC)} \\
Method & \rotatebox{0}{1\%} & \rotatebox{0}{10\%} & \rotatebox{0}{100\%}\\
\midrule[1.2pt]
Random & 58.1 & 69.1 & 79.0 \\
ImageNet & 63.5 & 72.6 & 80.4 \\
\midrule
ConVIRT \cite{convirt} & 64.9 & 77.1 & 80.8 \\
GLoRIA \cite{huang2021gloria} & 59.7 & 74.3 & 80.0 \\
BioViL \cite{biovil} & 57.9 & 72.7 & 80.0 \\
MedKLIP$^{\star}$ \cite{medklip} & 60.9 & 74.8 & 80.1 \\
KAD$^{\star}$ \cite{kad} & 78.7 & 80.7 & 82.5 \\
\textbf{IMITATE} & \textbf{80.2} & \textbf{82.2} & \textbf{83.9} \\
\bottomrule[1.2pt]
\end{tabular}
}
\label{tab: res finetune}
\end{table}

\subsubsection{Semantic Segmentation and Object Detection}
As demonstrated in Tab.~\ref{tab:seg res}, IMITATE consistently exhibits superior performance compared to all SOTA methods across various data fractions in all segmentation and detection tasks.
Remarkably, IMITATE achieves a Dice score of 70.5\% on RSNA segmentation with only 1\% data fine-tuning, surpassing the performance of all other SOTA methods fine-tuned on 100\% data except for MedKLIP~\cite{medklip}. Moreover, when fine-tuning with just 1\% of data, IMITATE achieves a 3.9\% mAP on the Object-CXR dataset. In contrast,other methods struggle to even touch a 1\% mAP under the same data fraction.

The remarkable enhancements achieved by IMITATE across diverse downstream tasks underscore the advantages of leveraging hierarchical alignment with structured medical reports during pre-training. This approach yields more informed and general medical representations that are better suited for a wide range of downstream applications beyond high-level tasks.

\begin{table*}[t!]
    \centering
    \caption{Results of semantic segmentation on SIIM and RSNA datasets and object detection on RSNA and Object-CXR datasets. The best results for each setting are highlighted in bold, and the '-' denotes mAP values smaller than 1\%.  Methods with $\star$ use extra annotated data for pre-training.
    }
    \scalebox{1.1}{
    \begin{tabular}{l c c c c c c c c c c c c}
     \toprule[1.2pt]
    &  \multicolumn{6}{c}{Semantic Segmentation (Dice)} & \multicolumn{6}{c}{Object Detection (mAP)}\\
     \midrule[1.2pt]
    & \multicolumn{3}{c}{SIIM} & \multicolumn{3}{c}{RSNA} & \multicolumn{3}{c}{RSNA} & \multicolumn{3}{c}{Object CXR}\\
    Method & 1\% & 10\% & 100\% & 1\% & 10\% & 100\% & 1\% & 10\% & 100\% & 1\% & 10\% & 100\% \\
    \midrule
    Random & 9.0 & 28.6 & 54.3 & 6.9 & 10.6 & 18.5  & 1.0 & 4.0 & 8.9 & - & 0.5 & 4.4\\
    ImageNet & 10.2 & 35.5 & 63.5 & 34.8 & 39.9 & 64.0 & 3.6 & 8.0 & 15.7 & - & 2.9 & 8.3 \\
   \midrule
    ConVIRT\cite{convirt} & 25.0 & 43.2 & 59.9 & 55.0 & 67.4 & 67.5 & 8.2 & 15.6 & 17.9 & - & 8.6 & 15.9 \\
    GLoRIA\cite{huang2021gloria} & 35.8 & 46.9 & 63.4 & 59.3 & 67.5 & 67.8 & 9.8 & 14.8 & 18.8 & - & 10.6 & 15.6 \\
    GLoRIA-MIMIC~\cite{huang2021gloria} & 37.4 & 57.1 & 64.0 & 60.3 & 68.7 & 68.3 & 11.6 & 16.1 & 24.8 & - & 8.90 & 16.6\\
    MGCA~\cite{mgca} & 49.7 & 59.3 & 64.2 & 63.0 & 68.3 & 69.8 & 12.9 & 16.8 & 24.9 & - & 12.1 & 19.2 \\ 
    MedKLIP$^{\star}$~\cite{wu2023medklip} & 50.2 & 60.8 & 63.9 & 66.2 & 69.4 & 71.9 & 8.9 & 16.3 & 24.5 & - & 7.1 & 11.6 \\ 
    \midrule
    \textbf{IMITATE} & \textbf{53.9} & \textbf{61.7} & \textbf{64.5} & \textbf{70.5} & \textbf{71.4} & \textbf{73.8} & \textbf{15.3} & \textbf{19.7} & \textbf{26.4} & \textbf{3.9} & \textbf{12.7} & \textbf{20.3}\\

    \bottomrule[1.2pt]
    \end{tabular}
   }
    \label{tab:seg res}
\end{table*}

\subsubsection{Zero-shot Image Classification}
In order to assess the efficacy of VLP in establishing connections between vision and language, we conduct zero-shot classification experiments on RSNA and SIIM datasets~\cite{rsna,siim}. The results for both datasets are presented in Tab.~\ref{zero-shot_class}. IMITATE surpasses other SOTA methods in three average metrics and performs well on both datasets. This outcome underscores the efficacy of IMITATE in the vision-language task.

\begin{table}[t]
\footnotesize
\centering
\caption{Comparison of various medical VLP techniques for zero-shot image classification tasks. Best results are emphasized in bold. Methods with $\star$ use extra annotated data for pre-training.}
\scalebox{0.85}{
\begin{tabular}{c|ccc|ccc}
\toprule[1.2pt]
Dataset & \multicolumn{3}{c|}{RSNA} & \multicolumn{3}{c}{SIIM}\\
Methods        & AUC(\%)      & F1(\%)       & ACC(\%)   & AUC(\%)      & F1(\%)       & ACC(\%) \\ \midrule[1.2pt]
ConVIRT~\cite{convirt}          & 80.4 & 58.4 & 76.1 & 64.3 & 43.3   & 57.0 \\
GLoRIA~\cite{huang2021gloria}           & 71.5 & 49.0 & 71.3   & 53.4 & 38.2 & 40.5 \\
BioViL~\cite{biovil}           & 82.8    & 58.3   & 76.7 & 70.8  & 48.6   & 69.1  \\
CheXzero$^{\star}$~\cite{chexzero} & 85.8 & 62.1 & 79.4 & 68.8 & 47.0 & 54.7 \\
MedKLIP$^{\star}$~\cite{medklip}             & 86.9    & 63.4 & 80.0 & 89.2    & 68.3 & 84.3 \\ 
\midrule
\textbf{IMITATE}             & \textbf{87.5}    & \textbf{64.6} & \textbf{81.3} & \textbf{89.9}    & \textbf{69.1} & \textbf{85.2} \\ 
\bottomrule[1.2pt]
\end{tabular}
}
\label{zero-shot_class}
\end{table}

\subsubsection{Image-text Retrieval}
To assess the alignment of image and text cross-modal features, we implemented the image-text retrieval task following GLoRIA \cite{huang2021gloria}. As other comparison methods do not include this task, we use the original result table from GLoRIA for a fair comparison. As shown in Table \ref{retrieval}, IMITATE outperforms both ConVIRT \cite{convirt} and GLoRIA \cite{huang2021gloria}, demonstrating that our method effectively aligns cross-modal features between images and text. IMITATE excels in cross-modal tasks.

\begin{table}[h]
\centering
\caption{
Results of image-text retrieval on the CheXpert $5 \times 200$ dataset are displayed, reporting top $K$ Precision metrics for $K=5, 10, 100$. The highest performances are highlighted in bold.
}
\label{retrieval}
\scalebox{0.9}{
\begin{tabular}{l|ccc}
\toprule[1.2pt]
 Method & Precision@ 5 & Precision@ 10 & Precision@ 100 \\
\midrule [1.2pt]
ConVIRT \cite{convirt} & 66.98 & 63.06 & 49.03 \\
GLoRIA \cite{huang2021gloria} & 69.24 & 67.22 & 53.78 \\
\textbf{IMITATE (ours)} & \textbf{71.83} & \textbf{69.75} & \textbf{55.21} \\
\bottomrule [1.2pt]
\end{tabular}
}
\end{table}

\subsection{Ablation Studies}
\subsubsection{Improvement of Hierarchical Alignment}
\label{sec: multi align}
We first investigate the effect of hierarchical alignment on pre-training. We conduct pre-training with various loss combinations. As indicated in Tab.~\ref{tab: ablation loss}, only aligning high-level features in V-V or V-L branch leads to poor performance, while IMITATE consistently improves results across all four datasets. 
Furthermore, the performance on four downstream tasks substantially drops when we only align multi-level visual features with both parts of the report in the `VLM' row. 
This demonstrates that aligning the two parts, which have different semantic levels in the report, with the same target will decrease the quality of the learned features, hence indicating that hierarchical alignment with the two parts individually is necessary.
Additionally, we observe that hierarchical alignment in the V-L branch produces better outcomes than hierarchical alignment in the V-V branch. This suggests that hierarchical alignment is more advantageous to vision-language than vision-vision contrasting. Combining all losses yields consistently superior results, indicating that joint hierarchical vision-vision and vision-language alignments are beneficial for VLP.

\begin{table}[h!]
\centering
\caption{Ablation study of each loss in IMITATE.
VVH: High-level alignment in V-V.
VLH: High-level alignment in V-L.
VVM: Multi-level alignment in V-V.
VLM: Multi-level alignment in V-L.
}
\label{tab: ablation loss}
\scalebox{1}{
\begin{tabular}{r|c|c|c|c}
\toprule
 & CheXpert & SIIM & RSNA & ChestXray-14\\
 & AUC(\%) & Dice(\%) & mAP(\%) & AUC(\%)\\
 & 1\% & 1\% & 1\% & 1\% \\ \midrule
VVH  & 87.5 & 32.1 & 7.6 & 64.5 \\
VVH,VLH  & 88.4 & 35.4 & 8.5 & 67.2 \\
{VLM}  & {84.2} & {30.4} & {5.7} & {61.3} \\
VVH,VLH,VVM  & 88.7 & 37.3 & 12.1 & 73.4 \\
VVH,VLH,VLM  & 88.8 & 38.2 & 12.6 & 74.6 \\
IMITATE  & \textbf{89.1} & \textbf{53.9} & \textbf{15.3} & \textbf{80.2}\\
\bottomrule
\end{tabular}
}
\end{table}

\subsubsection{Alignment with different Parts of Medical Reports}
\label{sec: diff report}
After showing the importance of hierarchical vision-language alignment, we experiment with structured reports to study the effect of report hierarchy on hierarchical alignment.
The results of IMITATE aligned with different parts of reports during pre-training are presented in Tab.~\ref{tab: align target}. 
Notably, pre-training IMITATE with hierarchical medical reports yields the highest performance across four tasks. This outcome underscores the efficacy of aligning with various report segments based on their inherent structure.

On the contrary, the lowest performance is observed when aligning the reversed target, implying the alignment of high-level visual features with the `Findings' section and middle-level features with the `Impressions' section. In contrast, aligning solely with the `Impression' section and concatenating `Findings' and `Impressions' as the vision-language alignment target do not show any significant improvements due to the ambiguity arising from the absence of the medical reports hierarchy. This is likely due to the ambiguity stemming from the absence of the hierarchical structure in medical reports.

\begin{table}[h!]
\centering
\caption{Ablation study of the different parts of reports for IMITATE.
Find\textbf{/}Imp indicates the `Findings' and `Impression' part of medical reports. Find\&Imp notes the concatenation of these two parts as one. `reversed' indicates switching two parts of reports for alignment.}
\label{tab: align target}
\scalebox{1}{
\begin{tabular}{r|c|c|c|c}
\toprule
& CheXpert & SIIM & RSNA & ChestXray-14\\
 & AUC(\%) & Dice(\%) & mAP(\%) & AUC(\%)\\
& 1\% & 1\% & 1\% & 1\% \\ \midrule
Imp   & 87.5 & 33.6& 12.2 & 77.5 \\
Find\&Imp   & 88.2 & 35.4 & 13.4  & 78.4 \\
reversed & 83.4 & 29.8 & 12.7  & 67.6 \\
IMITATE  & \textbf{89.1} & \textbf{53.9} & \textbf{15.3} & \textbf{80.2}\\
\bottomrule
\end{tabular}
}
\end{table}

\subsubsection{Impact of Clinical-Informed Contrastive Loss}
\label{abla: cicl}
In this section, we investigate the impact of $\mathcal{L}^{CICL}$ and different smooth kernels with $\lambda$ on the effectiveness of IMITATE. The outcomes are presented in Tab.~\ref{tab: smooth}.

\noindent\textbf{Clinical-Informed Contrastive Loss}\hspace{2mm}
We observe a significant reduction of performance when using $\mathcal{L}^{CL}$ instead of $\mathcal{L}^{CICL}$ in contrastive learning. This indicates that the clinical prior is a crucial component when using $\mathcal{L}^{CICL}$. 

\noindent\textbf{Smooth Kernel}\hspace{2mm}
Furthermore, we evaluate the results for various smooth kernels in Eq.~\eqref{eq:smooth} and find that the smoothed Exponential kernel outperforms the others, as shown in Tab.~\ref{tab: smooth}. The Gaussian and Laplacian kernels convert negative correlation coefficients to positive values, which can disrupt prior knowledge. The Sigmoid kernel preserves the coefficient range in $[-1,1]$ but may lead to strong penalization during pre-training, resulting in substandard performance. Gaussian, Laplacian, and Sigmoid kernels all exhibited poorer results than the smoothed Exponential kernel (Eq.~\eqref{eq:smooth}), which shrinks the coefficient range but does not convert negative values to positive ones, thereby preserving most prior knowledge and leading to superior performance.

\begin{figure*}[htp!]
\centering
\subfloat[CheXpert]{\includegraphics[width=0.32\linewidth]{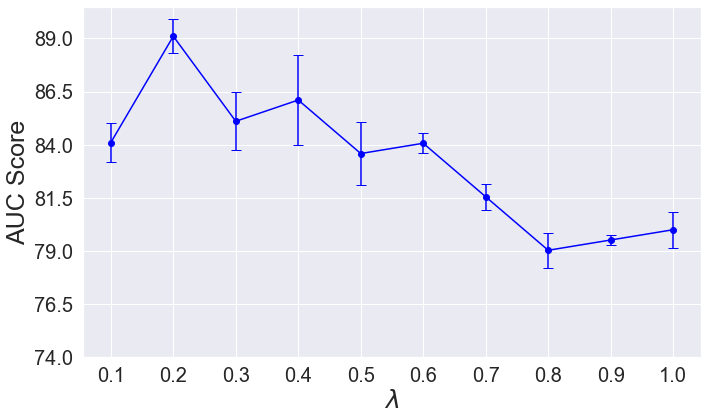}}
\subfloat[SIIM]{\includegraphics[width=0.32\linewidth]{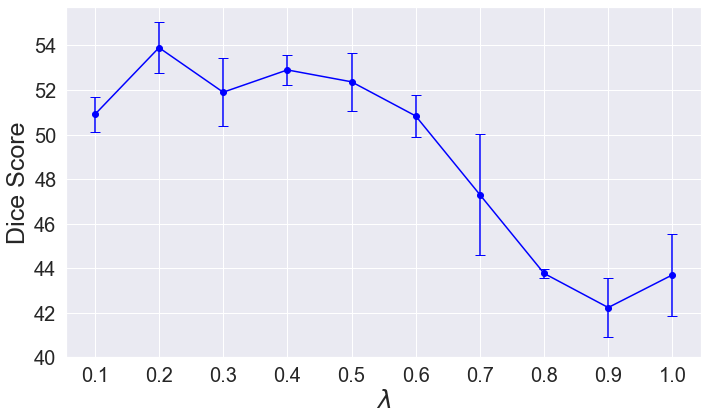}}
\subfloat[RSNA]{\includegraphics[width=0.32\linewidth]{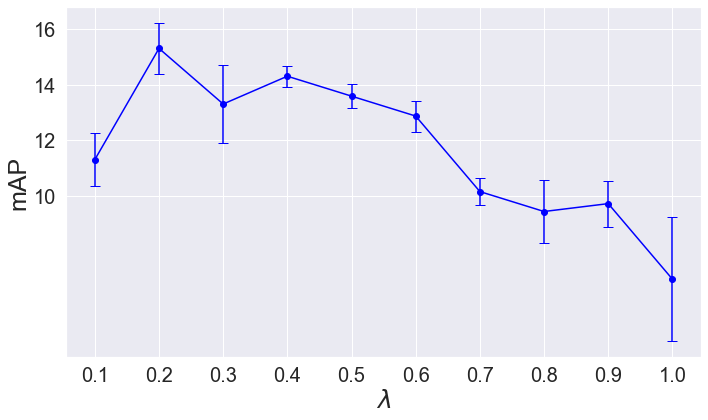}}\\
\caption{Performance of supervised image linear classification on CheXpert~\cite{irvin2019chexpert}, semantic segmentation on the SIIM~\cite{siim}, and object detection on the RSNA~\cite{rsna} datasets with 1\% labeled data fine-tuning, while varying $\lambda$ from $0.1$ to $1.0$.}
\label{fig: lambda analysis}
\end{figure*}

\noindent\textbf{Hyperparameter Sensitivities Analysis}\hspace{2mm} We evaluate the sensitivity of pre-training to different values of $\lambda$ on various downstream tasks. As shown in Fig.~\ref{fig: lambda analysis}, all pre-trained models with $\lambda \leq 0.4$ outperform the best baseline on three downstream tasks, while $\lambda \geq 0.5$ led to worse performance. It is crucial to note that excessive values of $\lambda$ can lead to bias due to the lack of control over the prior knowledge. Therefore, our framework's performance is stable for various downstream tasks when the strength of $\lambda$ is constrained to a small range. This finding suggests that the strength of clinical prior knowledge should be controlled within a certain range since the reports' correlation should only be considered as a weak constraint.

\begin{table}[h!]
\centering
\caption{Ablation of different smooth kernels. `w/o $\mathcal{L}^{CICL}$' indicates that only using the original contrastive loss~\cite{clip} as Eq.~\eqref{eq: ori cl loss} for VLP.}
\label{tab: smooth}
\scalebox{1}{
\begin{tabular}{r|c|c|c|c}
\toprule
 & CheXpert & SIIM & RSNA & ChestXray-14\\
 & AUC(\%) & Dice(\%) & mAP(\%) & AUC(\%)\\
 & 1\% & 1\% & 1\% & 1\% \\ \midrule
w/o $\mathcal{L}^{CICL}$ & 87.9 & 26.8 & 11.5 & 78.6 \\ \midrule
Gaussian  & 86.5 & 32.1 & 10.7 & 75.1 \\
Laplacian  & 87.1 & 31.7 & 11.4 & 75.7 \\
Sigmoid  & 85.7 & 30.2 & 10.9 & 75.2 \\

IMITATE  & \textbf{89.1} & \textbf{53.9} & \textbf{15.3} & \textbf{80.2}\\
\bottomrule
\end{tabular}
}
\end{table}

\noindent\textbf{IMITATE compared with unfrozen variants}\hspace{2mm}
Tab.~\ref{tab: ft cls},\ref{tab: res finetune},\ref{tab:seg res} present the performance of all downstream tasks when using IMITATE pre-trained with a frozen language model. In this section, we sequentially unfreeze the last six layers of the language model to evaluate the effectiveness of the trainable language model on downstream tasks for four datasets. The outcomes are reported in Tab.~\ref{tab:unfreeze}. We observed that the performance did not improve as we increased the number of unfrozen layers, while the training cost increased. This suggests that the trainable language model could be ablated to reduce training costs significantly. Furthermore, using a frozen language model can alleviate perturbations to visual feature learning from language embeddings.

\begin{table}[ht!]
\centering
\caption{Performance of unfrozen variants on downstream tasks. $\text{Unfreeze}_{n}$: the last $n$ layers of the language model are unfrozen.}
\scalebox{0.85}{
\begin{tabular}{r|c|c|c|c|c}
\toprule
 & Trainable & CheXpert & SIIM & RSNA & ChestXray-14\\
 & Parameters (M) & AUC(\%) & Dice(\%) & mAP(\%) & AUC(\%)\\
 & & 1\% & 1\% & 1\% & 1\% \\ \midrule
$\textrm{Unfreeze}_{1}$ & 59.1 & 87.2 & 45.3 & 13.7 & 78.7 \\
$\textrm{Unfreeze}_{2}$ & 66.1 & 87.8 & 50.5 & 14.6 & 77.9 \\
$\textrm{Unfreeze}_{3}$ & 73.2 & 88.5 & 47.4 & 14.8 & 78.4 \\
$\textrm{Unfreeze}_{4}$ & 80.3 & 88.2 & 50.9 & 15.1 & 79.6 \\
$\textrm{Unfreeze}_{5}$ & 87.4 & 88.8 & 45.7 & 14.2 & 78.1 \\
$\textrm{Unfreeze}_{6}$ & 94.4 & 88.4 & 37.5 & 14.3 & 78.5 \\ \midrule
IMITATE & \textbf{51.9}  & \textbf{89.1} & \textbf{53.9} & \textbf{15.3} & \textbf{80.2}\\
\bottomrule
\end{tabular}
}
\label{tab:unfreeze}
\end{table}

\noindent\textbf{Affect of Text Encoder}\hspace{2mm}
To demonstrate our method, IMITATE is text encoder agnostic. We conducted an ablation study with different text encoders to demonstrate that our work is text encoder agnostic. As shown in Table \ref{tab: abla text encoder}, the performance with three different text encoders on four downstream tasks does not show considerable fluctuation. This demonstrates that the frozen text encoder during VLP is versatile and is not limited to a specific text encoder.

\begin{table}[h!]
    \centering
    \caption{{Performance of various text encoder on downstream
tasks. Although different text encoders show varying performance fluctuations, these are marginal, demonstrating that our method, IMITATE, is not sensitive to the text encoder used.}}
    \scalebox{0.9}{
    \begin{tabular}{r|c|c|c|c}
    \toprule
     & CheXpert & SIIM & RSNA & ChestXray-14\\
     & AUC(\%) & Dice(\%) & mAP(\%) & AUC(\%)\\
     & 1\% & 1\% & 1\% & 1\% \\ 
     \midrule
    ClinicalBERT \cite{clinicalbert} & 88.7 & \textbf{54.1} & 14.9 & 80.0\\
    RadBERT \cite{radbert} & \textbf{89.3} & 53.7 & 15.0 & 79.8 \\
    BioClinicalBERT \cite{alsentzer2019publicly} & 89.1 & 53.9 & \textbf{15.3} & \textbf{80.2}\\
    \bottomrule
    \end{tabular}
    }
    
    \label{tab: abla text encoder}
\end{table}

{\noindent\textbf{Affect of Drop Ratio}\hspace{2mm}
We ablate the drop ratio with [0.25, 0.5, 0.75, 0.9] and report the results in Table \ref{tab: drop ratio}. The highest drop ratio, 0.9, provides the best performance, while other drop ratios show various negative effects. The decrease could be attributed to the redundant information from multi-level visual features, and a larger drop ratio can be treated as a strong latent augmentation to improve the learning quality.}

\begin{table}[ht!]
\centering
\caption{{The impact of different drop ratios on the performance of various downstream tasks. This indicates that performance increases with a higher drop ratio.}}
\label{tab: drop ratio}
\scalebox{0.8}{
\begin{tabular}{c|c|c|c|c}
 \toprule
 Drop Ratio & CheXpert & SIIM & RSNA & ChestXray-14\\
 & AUC(\%) & Dice(\%) & mAP(\%) & AUC(\%)\\
 & 1\% & 1\% & 1\% & 1\% \\ \midrule
0.25  & 85.8 & 49.5 & 10.9 &  77.5 \\
0.5   & 87.2 & 50.1 & 11.6 &  77.8 \\
0.75   & 88.5 & 52.6 & 13.8 &  78.9 \\
\textbf{0.9 (ours)}  & \textbf{89.1} & \textbf{53.9} & \textbf{15.3} & \textbf{80.2}\\
\bottomrule
\end{tabular}
}
\end{table}

\subsection{Visualizing Qualitative Results}
\begin{figure}[h!]
    \centering
    \includegraphics[width=1\linewidth]{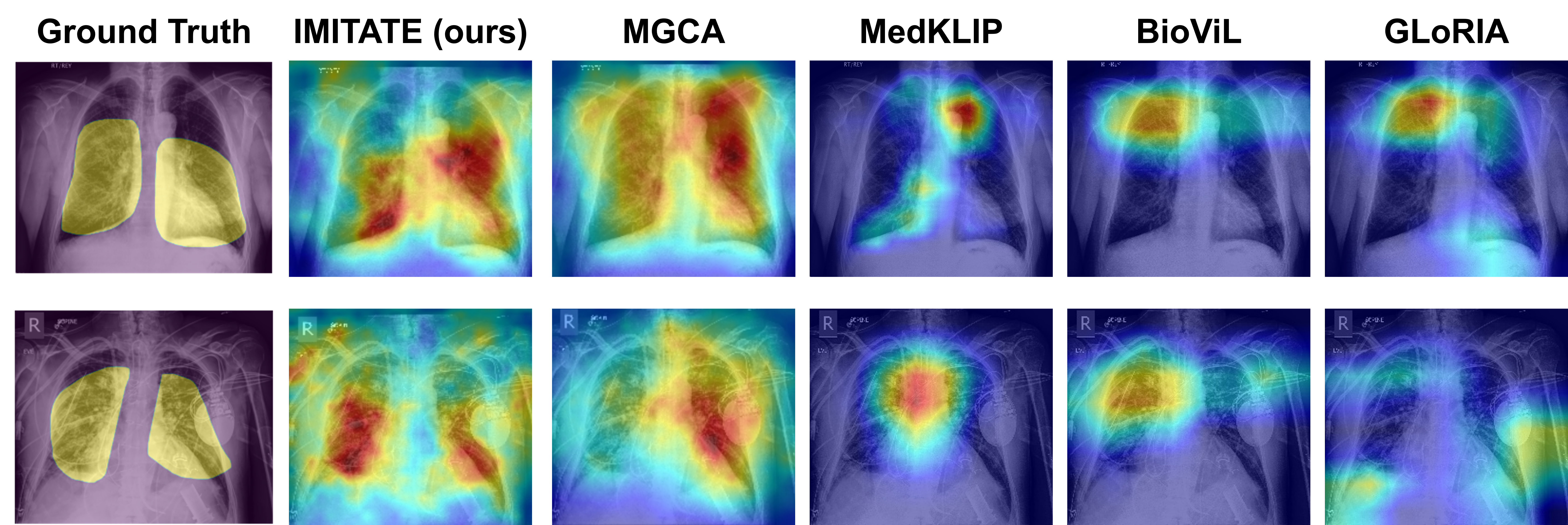}
    \caption{{Comparison between the saliency map generated by the vision encoder after VLP and the region of interest identified by radiologists.}}
    \label{fig: sal vis}
\end{figure}

To delve deeper into the learned visual knowledge from IMITATE, we utilized Grad-CAM~\cite{selvaraju2017grad} to produce saliency maps for CXR images derived from the model in its pre-trained state. We select two representative CXR images showcasing two prevalent diseases, \textit{Edema} and \textit{Lung Opacity}. Notably, each of these images comes with ground truth annotation pinpointing the region of concern, as documented in \cite{chex500}. As evident from Fig.~\ref{fig: sal vis}, IMITATE boasts an impressive capability to accurately delineate the clinical regions of concern in the CXR images, {outperforming other compared methods \cite{mgca,medklip,kad,biovil,huang2021gloria}}. This is particularly noteworthy considering that IMITATE achieves this precision without relying on any external prompts or the need for additional model fine-tuning.

\section{Conclusion and Discussion}

This study introduces a novel VLP framework that imitates the human understanding of paired image-text in a hierarchical manner.
This framework, named IMITATE, aligns CXR images and hierarchical medical reports at multiple levels. IMITATE utilizes hierarchical alignment and different parts of medical reports to enhance image representation by incorporating hierarchical information from medical report structures. Notably, this operation requires no additional data or manual pre-processing. Moreover, we propose Clinical-Informed Contrastive Loss, which explicitly integrates clinical prior knowledge through smoothed medical report correlation.

To best of our knowledge, IMITATE is the first framework to align hierarchical information from structured medical reports to multi-level visual features in medical images. Furthermore, we incorporate the clinical similarity into the contrastive loss. These contributions address a critical limitation in existing VLP approaches that ignore clinical similarity among patients. Furthermore, hierarchical alignment and $\mathcal{L}^{CICL}$ provide more reasonable learning targets for the visual modality, resulting in significant improvements in the performances of all downstream tasks with a 50\% reduction in trainable parameters compared to other SOTA methods.
We believe that this framework will benefit the medical domain, as hierarchical medical report generation is a standard procedure without extra cost. Additionally, it will inspire the general VLP domain, as hierarchical information commonly exists worldwide, such as title and content, caption, and description, among others.
\bibliographystyle{IEEEtran}
\bibliography{mybib}
\end{document}